\documentclass[twocolumn,times,final,authoryear]{elsarticle}

\usepackage{prletters}
\usepackage{framed,multirow}
\usepackage{soul}
\usepackage{dingbat}
\usepackage{graphicx}
\usepackage{xcolor}
\usepackage{caption}
\usepackage{subcaption}
\usepackage{amsmath}

\usepackage{amssymb}
\usepackage{latexsym}

\usepackage{url}
\usepackage{xcolor}
\definecolor{newcolor}{rgb}{.8,.349,.1}

\begin{document}

\begin{frontmatter}

\setcounter{page}{1}

\title{Long-Tail Learning with Rebalanced Contrastive Loss}

\author[1]{Charika \surname{De Alvis}\corref{cor1}} 
\cortext[cor1]{Corresponding author: 
 }
\ead{charika.weerasiriwardhane@sydney.edu.au}
\author[1]{Dishanika \surname{Denipitiyage}}
\ead{Dishanika.Denipitiyage@sydney.edu.au}
\author[1]{Suranga \surname{Seneviratne}}
\ead{Suranga.Seeviratne@sydney.edu.au}

\affiliation[1]{organization={The University of Sydney},
                addressline={Camperdown NSW }, 
                city={Sydney}, 
                postcode={2050}, 
                state={NSW},
                country={Australia}}

\begin{abstract}
Integrating supervised contrastive loss to cross entropy-based classification has recently been proposed as a solution to address the long-tail learning problem. However, when the class imbalance ratio
 is high, it requires adjusting the supervised contrastive loss to support the tail classes, as the conventional contrastive learning is biased towards head classes by default. To this end, we present Rebalanced Contrastive Learning (RCL), an efficient means to increase the long-tail classification accuracy by addressing three main aspects: 1. Feature space balancedness -- Equal division of the feature space among all the classes 2. Intra-Class compactness -- Reducing the distance between same-class embeddings 3. Regularization -- Enforcing larger margins for tail classes to reduce overfitting. RCL adopts class frequency-based SoftMax loss balancing to supervised contrastive learning loss and exploits scalar multiplied features fed to the contrastive learning loss to enforce compactness. We implement RCL on the Balanced Contrastive Learning (BCL) Framework, which has the SOTA performance. Our experiments on three benchmark datasets demonstrate the richness of the learnt embeddings and increased top-1 balanced accuracy RCL provides to the BCL framework. We further demonstrate that the performance of RCL as a standalone loss also achieves state-of-the-art level accuracy. 
\end{abstract}

\end{frontmatter}


\section{Introduction}
\label{sec1}
The data imbalance between different classes usually hinders training an unbiased classifier. This is a significant problem, especially when the tail class is of great importance for applications such as pedestrian classification, fraud detection, and spam filtering. While approaches such as data resampling and loss reweighting are used to alleviate class imbalance, these approaches increase the tail classes' accuracy by sacrificing the head class accuracy.
An important aspect of a robust model is the capability to address the distribution shift from training (long-tail) to unknown test (uniform or inversely long-tailed) distributions. Later in some frameworks, cross-entropy loss for long-tail learning is combined with metric learning or contrastive learning. According to \cite{elm} 
strong performance improvement occurs when the cross-entropy loss is combined with Supervised Contrastive Learning (SCL). This work demonstrates that learning better embeddings can assist in resolving most of the long-tail classification challenges, as better embeddings can contribute to more generalized classifiers. Along this line, one approach is decoupling representation and classifier learning (\cite{kang2019decoupling}). However, the representations learned through SCL (\cite{scl}) cannot fully eliminate the dominance of the most frequent classes or maintain the linear separability of embeddings from head-to-tail classes without necessary adjustments for long-tail distributions. Therefore, improving SCL to handle long-tail data distributions more robustly is important.

\cite{bcl} propose balanced contrastive loss BCL, an improved SCL loss for the context of long-tail learning where the main intuition is to average the attraction and repulsion terms based on the mini-batch class frequency so that each class equally contributes to the optimization process. We propose a means to further enhance the quality of the learnt embedding of BCL from the perspectives of balancedness, intra-class compactness, and regularization using Rebalanced Contrastive Learning (RCL). We emphasize that the RCL is not limited to working with BCL only. Given its simplicity of implementation, it can be incorporated into any SCL method. More specifically, we make the following contributions.
\begin{itemize}
\item We propose  RCL  to optimize the learnt representation of  SCL/BCL through i) 
 establishing the robustness for distribution shift from train (long-tail) to test (uniform) using the adopted class frequency-based SoftMax balancing, ii) enforcing intra-class compactness for tail classes through scalar multiplied features fed to the contrastive loss, and iii) enforcing better class separation through enforced class frequency based embedding margins.
\item By conducting experiments on the long-tail versions of the three benchmark datasets: CIFAR10\_Lt, CIFAR100\_Lt, and ImageNet\_Lt, we show that the addition of RCL over SCL/BCL improves the top-1 balanced accuracy. Our ablation study further shows that RCL as a standalone metric also increases the top-1 harmonic mean accuracy, indicating improvements in the least-performing classes.
   
 \item We conduct qualitative and quantitative evaluations of the learnt embeddings and show that adding RCL to BCL enhances class separability with larger margins through balancedness, intra-class compactness, and regularization.
 
\end{itemize}

\section{Related Works}
 We explain related work in long-tail classification methods under three topics: i) Data-level solutions, ii) Algorithmic-level solutions, and iii) SCL-based approaches.  
 
\subsection{ Data-Level Solutions}Early works in long-tail learning include data-level solutions such as random oversampling, random undersampling (\cite{126}), 
square root sampling (\cite{root}) and progressively balanced sampling (\cite{kang2019decoupling}) that suffer from over-fitting or under-fitting to data. Re-weighting methods (\cite{cb,shu2019meta}) assign weights on different training samples based on the label or instance (hard or easy samples) and have shown significant accuracy increments over few-shot classes. However, re-weighting methods tend to sacrifice the many-shot accuracy. Classifier balancing techniques such as fine-tuning the learnt classifier (\cite{guerriero2018deepncm}), post-hoc calibrations (\cite{menon2020long,kang2019decoupling}), domainshift calibrations (\cite{peng2022optimal}) alleviate the imbalances of the data. However, these methods may not be sufficient under extreme class imbalance factor and when the number of classes is very high. Data augmentation is also used to improve accuracy for classes suffering from absolute rarity (\cite{150}). \cite{102} use generative models to generate new samples for tail classes as a convex combination of existing samples. Transferring the information from head classes to tail classes also provides promising results (\cite{100}). Yet, these methods can not completely eliminate the tail class over-fitting and the dominance of head classes. 

\subsection{Algorithmic-Level Solutions}
Under algorithmic level solutions integrating a per-class margin (\cite{menon2020long,metasoftmax}) into the cross-entropy loss is a popular work direction. In most cases, the logit margin is set as inversely proportional to the class frequency to enforce a larger margin for tail classes to enhance generalization. However, these methods do not control the learned feature distributions explicitly. Therefore, it is important to use techniques to optimize the learned feature distribution to avoid the diffusion of the rare classes. 

\subsection{Representation Optimization With SCL }The contrastive learning-based methods have been used recently in the context of long-tail learning as they contribute to better feature representation by aligning the same class pairs together and repulsing the different class pairs. \cite{target} propose a novel technique to improve SCL to compute optimal class centres and evenly balance the features space. Then, they associate the generated class centres with actual class IDs, considering mutual closeness. However, the results here could be suboptimal since there is no deterministic strategy to associate the class centres with the true classes. \cite{prototype} introduce an improved SCL framework denoted as prototypical supervised contrastive (PSC) learning where it learns prototypes for each class; subsequently, each instance is to be pushed towards the prototypes of its corresponding class and pulled away from the other class prototypes. In this approach, tail class instances may have a larger effect on the model optimization than head class instances, sacrificing the head class accuracy. To address this issue, \cite{sub} showcase subclass-balancing contrastive learning where the loss is defined on subclasses equal to the tail class size inside the head classes through clustering. Here, the authors balance the  SCL loss accordingly to support long-tail classification scenarios with higher class imbalance and subsequently balanced SCL loss is jointly minimized with CE loss. Our work focus on a novel yet simpler approach based on BalanceSoftmax concept to efficiently balance the SCL component so that it can significantly enhance the balancedness, intra-class compactness, and tail-class margins through rebalancing the BCL loss. 
\section{Methodology}
\label{sec:formatting}
\subsection{Notations}
   Multiclass classification problem can be defined by input instance domain $\mathcal{X}$, class label domain is given by $ Y=[L]=\{1,……,L\}$. For a set of training samples $ S = {(x_i,y_i)}_{i=1}^N \sim^{i.i.d} P^N  $ where joint distribution P characterized on $\mathcal{X} \times Y$. All the weights corresponding to the model are denoted by $W$. The multi-class learning task is to learn a scorer $f : \mathcal{X} \to R^L$ so that it minimizes the expected loss $l:Y \times R^L \to R_+$.
The scorer corresponding to class y is $f_y(x_i) = w_y^T z_i +b_y$, where $ w_y \in R^K$ are the weights of the last fully connected classification layer corresponding to class y. And $b_y \in R $ is the bias. Learned embedding corresponding to the input $x_i$ is $z_i \in R^K$. Further, we denote the batch frequency of class y by $|By|$ and the learnt class prototype by $c_y$ as defined in the work of \cite{bcl}.

\subsection{Preliminaries}
\label{l1}
\subsubsection{ CE Based Classification:} Our work makes use of fundamental Cross Entropy (CE) modification methods for long-tail learning. Firstly, as the SoftMax function is inherently biased under long-tail learning, a Balanced SoftMax function (\cite{metasoftmax}) is proposed to address the train-to-test distribution shift from a probabilistic perspective. The modified loss is denoted in Equation~\ref{2}. The authors provided an analytical derivation that optimizing for the Balanced SoftMax cross-entropy loss is essentially equivalent to minimizing the bound on generalization error. The second interesting approach is logit adjustment (LA) proposed by \cite{menon2020long} that allows larger margins for tail classes, which can be denoted with Equation~\ref{4}. Here $\pi_y$ denotes the probability of class y. This model is more favourable as $\tau$ provides the degree of freedom to adjust for different datasets and class imbalance levels.

\begin{equation}
\text{Balanced\_SoftMax\_Function} = - \log \frac{n_y e^{f_y(x)}}{\sum_{i \in L} n_i e^{f_i(x)}}
\label{2}
\end{equation}

\begin{equation}
\text{Logit\_Adjusted\_Loss}= - \log \frac{ e^{f_y(x)+\tau\log{\pi_y}}}{\sum_{i \in L}  e^{f_i(x)+\tau\log{\pi_i}}}
\label{4}
\end{equation}

 \subsubsection{ SCL Integration:} Above LA cross entropy classification module combined with a supervised contrastive learning module to enhance the long-tail classification accuracy. The supervised contrastive learning component is balanced through class averaging and class complements (\cite{bcl}). The new loss is denoted as balanced contrastive loss BCL. This loss avoids the bias occurred towards the head classes. Class complements is an effort to have all classes to be present in every mini batch through the use of class centre prototypes. When the rare class instances are missing from a batch the centre embedding would act as an actual training instance. Refer to the BCL loss component given in Equation~\ref{bcl} for the use 
of class averaging and class complements.
\begin{equation}
L_{\text{BCL}}=-\frac{1}{|B_y|} \cdot \\ \sum\limits_{p\in {B_y \{i\}} \cup \{c_y\}}\log \frac{e^{z_i.z_p}}{\sum\limits_{j\in Y}\frac{1}{|B_j|+1}\sum\limits_{k \in B_j \cup \{c_j\}} e^{z_i.z_k}}
\label{bcl}
\end{equation}

 Our proposed approach RCL is implemented on  BCL to incorporate more balancedness through the adopted balanced SoftMax concept and enforce larger margins for tail classes. The BCL + RCL  is integrated with the logit-adjusted CE loss to conduct the classification. Furthermore, RCL loss can be implemented on different supervised contrastive losses along with any combination of cross-entropy loss with class imbalance adjustment.

\begin{figure*}[t]
\centering
\hspace{-10mm}
\begin{subfigure}[]{0.45\textwidth}
\includegraphics[width=1\linewidth]{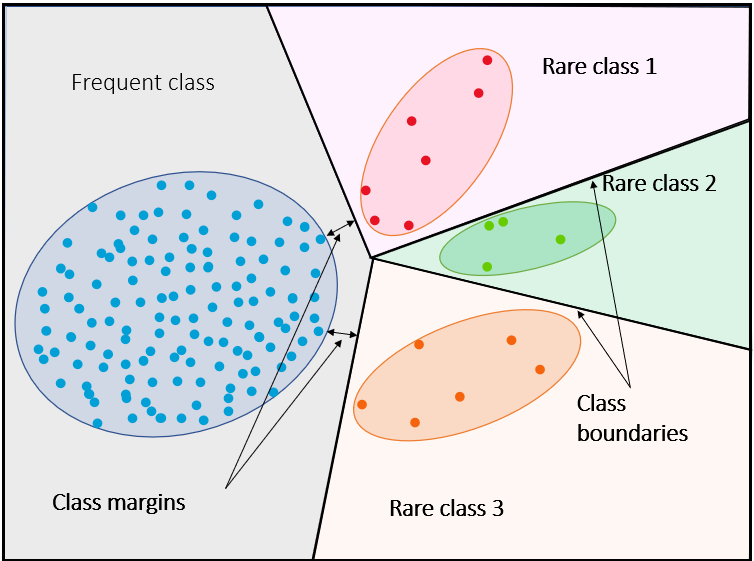}
\caption{ Learnt Feature Distribution }
 \label{cc1}
\end{subfigure}
\begin{subfigure}[]{0.45\textwidth}
\includegraphics[width=1.15\linewidth]{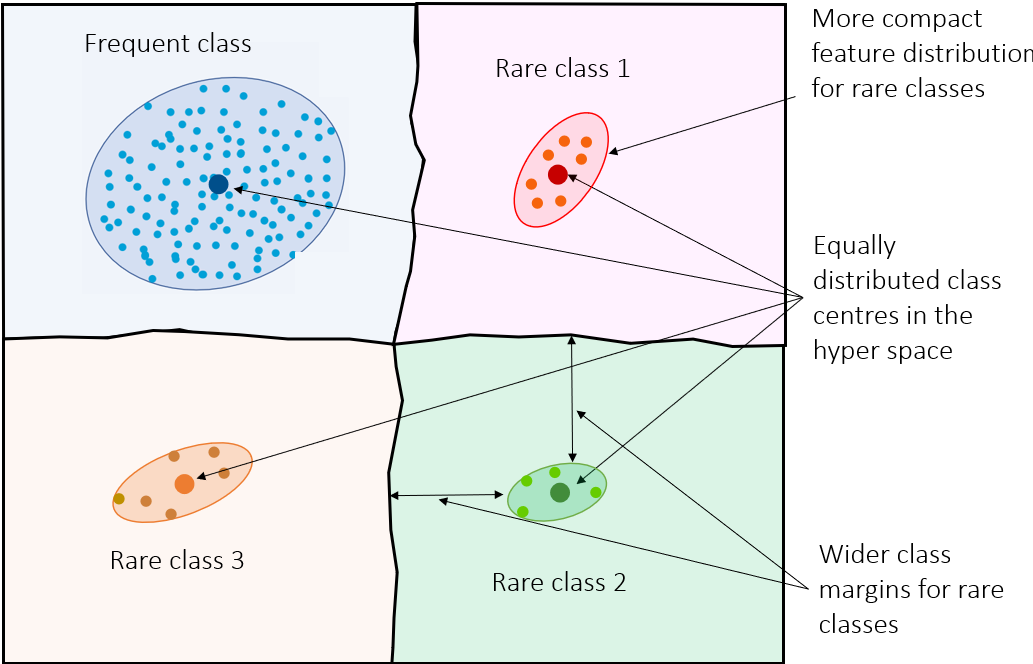}
\caption{Ideal Feature Distribution }
\label{cc3}
\end{subfigure}

\caption{ (a) This figure denotes the distribution of the penultimate layer embeddings in a 2D space in the class imbalanced 
case. Important factors to note are 1) feature distribution in the hyperspace, 2) class margins, and 3) feature deviation in each 
class. (b) Shows the expected ideal feature distribution under class imbalance. Symmetrically distributed class 
centres in the hyperspace, feature compactness, clear separability and generalisation of the class margins need to be addressed by the SCL components.}

\vspace{-5mm}
\end{figure*}
\begin{figure}[bt]
\centering
  \includegraphics[width=1.1\linewidth]{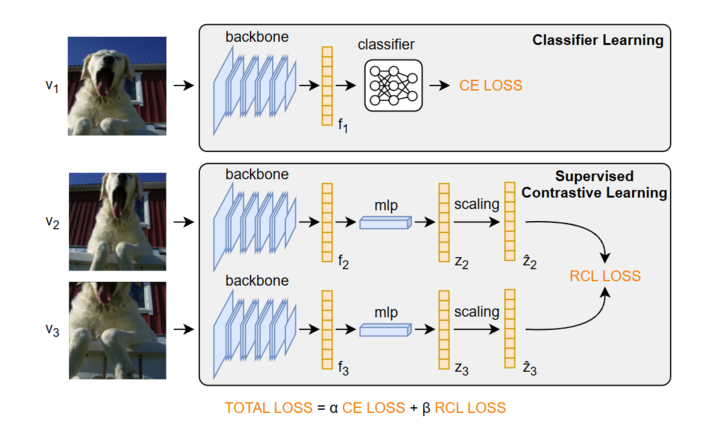}
\caption{RCL can be implemented with parallel branches as in the work of \cite{bcl} for classifier learning and supervised contrastive learning, as indicated in the figure. $v_1$ is the original image and $v_2$ and $v_3$ are augmentations of it. We adopt the same augmentation methods in BCL framework. A common backbone is used for both branches. Total loss is minimized during the training process.}
   \label{pipe}
   \vspace{-5mm}
\end{figure}
\subsection{Feature Representation Optimization With RCL}
Our RCL-based approach addresses three key aspects in the embedding space that can directly contribute to increasing accuracy: 1) Balanced feature space, 2) Intra-class compactness, and 3) Assigning larger margins to ambiguous classes. These key aspects are expected to provide more support for tail class classification. Figure~\ref{cc1} denotes the distribution of the learned embedding under the class imbalance for a simple separable 2D problem. The majority class tend to occupy the larger area of the feature space with larger margins, thus allowing larger generalisation on the head classes and over-fitting to minority classes. In Figure~\ref{pipe}, we illustrate the pipeline for the RCL-based learning. In this work, we analyse the RCL directly applied to the SCL loss, and secondly, RCL is applied to the BCL loss. 
\subsubsection{Balancing Feature Space}
It was noted that unsupervised contrastive learning methods are more robust to highly imbalanced class distributions compared to supervised contrastive learning (\cite{self}). The reason for this phenomenon is that the unsupervised CL methods generate a balanced feature space for all the classes compared to SCL. To this end, some research focuses on pushing the learnt embeddings into pre-defined class centres to obtain the balance (\cite{target}). In this work, we try to obtain balancedness by balancing the  SCL loss sufficiently. The modified SCL loss is defined as Rebalanced Contrastive Loss (RCL). We adopt the Balanced SoftMax concept described in Section~\ref{l1} to modify the SCL loss as indicated in Equation~\ref{rcl1}. Where $n_y$ is the frequency of the class y in the complete training set. 
\begin{equation}
L_{\text{RCL}}=-\frac{1}{|B_y|} \cdot \\ \sum\limits_{p\in {B_y \{i\}} }\log \frac{\mathcolor{blue}{n_y}e^{z_i.z_p}}{\sum\limits_{j\in Y} \mathcolor{blue}{n_j}\sum\limits_{k \in B_j } e^{z_i.z_k}}
\label{rcl1}
\end{equation}
\subsubsection{Enforcing Feature Compactness}
Minimizing the intra-class distance among the same class embeddings is a main aspect of interest as it reduces the deviation of the features across the decision boundary and enhances class discriminativeness. Feature Clusters Compression (FCC) (\cite{fcc}) is applied previously to decrease the diffusion of backbone features by compressing learnt embeddings. This is achieved by multiplying original learnt features by a selected scaling factor and using the multiplied
features in computing the cross entropy loss for classifier learning. In this work, we modify this approach and integrate the feature compression component directly into the RCL instead of sending it through the classifier as in CE minimization, i.e., $\tilde{z_i}=z_i * \tau_i$. The features are scalar multiplied before feeding to the SCL loss. The intuition is to provide a higher degree of freedom to reshape features based on feature compression. The scaling factor $\tau>0$ and value is selected based on the class id. At the $ epoch= \frac{total\_epochs}{2}$, the validation accuracy is obtained, and if the validation per class accuracy is lower than $(20\%)$, a novel scaling factor is applied for the corresponding classes and for the rest of the classes, the factor is set to 1 (only two $\tau$ values are used for the entire dataset). The scaling factor used on the underperforming classes is selected based on a grid search. The RCL modification with feature compression can be implemented on BCL to enhance the balancedness as in Equation \ref{bcl1}. The fraction in red is the class averaged similarity scores. 
\begin{equation}
L_{\text{BCL+RCL}}=-\frac{1}{|B_y|} \cdot \\ \sum\limits_{p\in {B_y \{i\}} \cup \{c_y\}}\log \frac{\mathcolor{blue}{n_y}e^{\tilde{z}_i.\tilde{z}_p}}{\sum\limits_{j\in Y} \mathcolor{blue}{n_j}\mathcolor{red}{\frac{1}{|B_j|+1}\sum\limits_{k \in B_j \cup \{c_j\}} e^{\tilde{z}_i.\tilde{z}_k}}}
\label{bcl1}
\end{equation}

\subsubsection{Regularization}
The rare classes must be more strongly regularized compared to frequent classes to reduce the generalization error of the rare classes without sacrificing the accuracy of majority classes. To achieve this it requires a data-dependant or label-dependent regularizer rather than the standard L2 regularize, which only depends on the model weights. Therefore, the data-dependent properties like class margins can be useful.  Enforcing a larger margin for a class can be interpreted as regularization (\cite{cao2019learning}) because the inverse of the minimum margin among all the examples is linked with the standard generalization error bounds. \cite{elm} demonstrate that enforcing logit margins is insufficient under extreme imbalance scenarios, and therefore, embedding margins are required to enhance the regularization of the minority classes. We demonstrate the RCL's importance in enforced embedding margins by reformulating it as in Equation~\ref{eq3} followed by Equation~\ref{eq4}. According to this format, it is evident that margin is implicitly enforced for the similarity score between the same class and dissimilar class embeddings based on the class frequency. We can further analyse the loss to get a comprehensive insight considering the pairwise margin loss.
\begin{equation}
L_{\text{RCL}}=-\frac{1}{|B_y|} \cdot \\ \sum\limits_{p\in {B_y \text{\textbackslash} \{i\}} }\log \frac{e^{log(n_y)}e^{\Tilde{z_i}.\Tilde{z_p}}}{\sum\limits_{j\in Y} \sum\limits_{k \in B_j } e^{log(n_j)}e^{\Tilde{z_i}.\Tilde{z_k}}}
\label{eq3}
\end{equation}
\begin{equation}
L_{\text{RCL}}=\frac{1}{|B_y|} \cdot  \sum\limits_{p\in {B_y \text{\textbackslash}\{i\}} }\log[1+ \sum\limits_{j\in Y} \sum\limits_{k \in B_j \text{\textbackslash} \{p\}}e^{\log (\frac{n_j}{n_y})}e^{\Tilde{z_i}.\Tilde{z_k} -\Tilde{z_i}.\Tilde{z_p}}]
\label{eq4}
\end{equation}

Pairwise margins $\log(\frac{n_j}{n_y})$ denoted in Equation \ref{eq4} enforces the desired gap between the similarity scores $\Tilde{z_i}.\Tilde{z_p}$ and $\Tilde{z_i}.\Tilde{z_k}$ depending on the belonginess to same or different classes. This imposes a larger margin between the tail intra-class similarity score and tail and head inter-class similarity score $(n_y<<n_j)$  so that it enhances the feature compactness within the tail class and increases the distance between head and tail feature clusters.  
Figure~\ref{cc3} denotes a 2D classification problem which portrays the ideally expected distribution of the learnt features under an imbalanced data distribution. All the classes occupy a similar area in the feature space, and the features are well compressed towards the class centre with larger margins for the rare classes.
\section{Experimental Setup and Results}

\begin{table}[h]
\centering
\caption{Ablation study for CIFAR100\_Lt with IF-100  to test the additive effects of representation learning with SCL, BCL, and RCL. The arithmetic mean of top-1 accuracy and the harmonic mean of top-1 accuracy are reported.\vspace{2mm}}
\begin{tabular}{ cccccc }
\hline
\textbf{LC} & \textbf{SCL} & \textbf{BCL} &\textbf{RCL} &\shortstack{\textbf{Arithmetic}} & \shortstack{\textbf{Harmonic}} \\ 
\hline
\checkmark&-&-&-&44.0&35.2\\

\checkmark&\checkmark&-&-&51.1&39.2\\
\checkmark&\checkmark&\checkmark&-&51.9&40.2\\

\checkmark&\checkmark&-&\checkmark&\textbf{52.2}&\textbf{42.4}\\
\checkmark&\checkmark&\checkmark&\checkmark&52.1&41.1\\
\hline
\end{tabular}
\label{t1}
\end{table}

\begin{table}[bt]
\centering
\caption{Ablation study for CIFAR10\_Lt with IF-100  to test the additive effects of representation learning with SCL, BCL, and RCL. The arithmetic mean of top-1 accuracy and the harmonic mean of top-1 accuracy are reported. \vspace{2mm}}
\begin{tabular}{ cccccc }
\hline
\textbf{LC} & \textbf{SCL}  & \textbf{BCL} &\textbf{RCL} &\shortstack{\textbf{Arithmetic}} & \shortstack{\textbf{Harmonic}} \\ 
\hline
\checkmark&-&-&-&81.2&80.5\\

\checkmark&\checkmark&-&-&83.5&83.1\\
\checkmark&\checkmark&\checkmark&-&84.3&83.7\\

\checkmark&\checkmark&-&\checkmark&86.0&85.3\\
\checkmark&\checkmark&\checkmark&\checkmark&\textbf{86.2}&\textbf{85.7}\\
\hline
\end{tabular}
\label{t2}
\vspace{-4mm}
\end{table}
\begin{table*}[bt]
\caption{Performance comparison table for RCL with SOTA methods for CIFAR100\_Lt (trained for 200 epochs and for class imbalance factor = 100). `-' is used when the SOTA results are not available in the literature. \vspace{2mm} }
\centering
\label{3}
\begin{tabular}{ ccccccc }
\hline
\textbf{Dataset} & \multicolumn{3}{c}{\shortstack{\textbf{CIFAR10\_Lt}}}& \multicolumn{3}{c}{\shortstack{\textbf{CIFAR100\_Lt}}} \\
\hline
Imbalance Ratio & 100&50&10 & 100&50&10  \\
\hline

Cross Entropy (CE)&70.4&74.8&86.4&38.5&43.9& 55.7 \\

LDAM+DRW (\cite{cao2019learning}) &77.2& 81.0&88.2  &42.2&46.6& 58.7 \\

LogitAdjust (\cite{menon2020long}) &80.2&-&-&43.9&-&57.4\\

CB+Focal (\cite{cb})&74.6& 79.4&87.1 &39.9&45.2&58.0 \\

BALMS (\cite{metasoftmax})&84.9&-& 91.3 &50.8& -&63.0 \\

\hline
\multicolumn{7}{c}{Two Stage Methods}\\
\hline
Decoupled (\cite{kang2019decoupling})&78.5&-&91.1&41.3&-&63.4\\

DRO\_LT (\cite{samuel2021distributional})&80.5&-&-&47.1&57.6&63.4 \\
\hline
\multicolumn{7}{c}{Ensemble Methods}\\
\hline
RIDE (\cite{ride})&-&-&-&49.1&-&61.8\\
ACE (\cite{ace})&81.4&84.9&-&49.6&51.9&-\\
\hline

\multicolumn{7}{c}{CE+ Other Losses}\\
\hline
BBN (\cite{zhou2020bbn})&79.8&82.2&-&42.6&47.0&-\\
\hline
\multicolumn{7}{c}{CE+Supervised Contrastive Loss}\\
\hline
Hybrid\_spc (\cite{proto})&81.4&83.7&91.1&46.7&48.9& 63.1\\

TSC (\cite{target})&79.7&82.9&88.7&43.8&47.4&59.0\\
BCL&84.3&87.2&91.1&51.9&56.6&64.9 \\

RCL (ours)&86.0&\textbf{88.5}&91.7&\textbf{52.2}&\textbf{56.8}&\textbf{65.6}\\

BCL+RCL (ours)&\textbf{86.2}&88.4&\textbf{92.0}&52.1&56.0&65.4\\
\hline

\end{tabular}
\end{table*}

\begin{table}[bt]
\caption{Performance comparison table for RCL with SOTA methods for ImageNet\_Lt for backbones Resnet50 and ResNext50 (trained for 90 epochs and class imbalance factor = 256 with sim-sim data augmentation) \vspace{2mm} }
\centering
\label{3}
\begin{tabular}{ ccc }
\hline
\textbf{Dataset}& \textbf{ResNet50}&\textbf{ResNext50} \\
\hline

CE&41.6&44.4\\

LDAM+DR  &49.8&-\\

LogitAdjust&50.9&-\\

CB+Focal&46.8&45.3\\

BalSoftMax&-&51.4\\

\hline
\multicolumn{3}{c}{Two Stage Methods}\\
\hline
Decoupled&50.6&49.6\\

DRO\_LT&53.5&53.5\\
\hline
\multicolumn{3}{c}{Ensemble Methods}\\
\hline
RIDE&55.4&56.8\\
ACE&54.7&56.6\\
\hline

\multicolumn{3}{c}{CE+Other Losses}\\
\hline
ELM&50.6&-\\
\hline
\multicolumn{3}{c}{CE+Supervised Contrastive Loss}\\
\hline
BBN&48.3&49.3\\

BCL &56.2&57.4\\
\hline

BCL+RCL (ours)&\textbf{56.3}&\textbf{57.5}\\
\hline

\end{tabular}
\label{imagenet}
\vspace{-5mm}
\end{table}

To validate our findings on representation learning optimization with RCL, we use three benchmark datasets with significant class imbalance, exhibiting long-tail distribution and involving a large number of classes. Next, we describe the datasets, experiment design followed by comparisons of the results with the SOTA and ablation study. 

\subsection{Datasets}
We use the long-tail versions of the benchmark datasets: CIFAR10\_Lt, CIFAR100\_Lt, 
and ImageNet\_Lt. The level of class imbalance is measured by the Class Imbalance Factor (IF), which is the ratio between the most frequent class frequency and the least frequent class frequency. We set the maximum possible imbalance factors for the datasets, leaving at least 5-6 instances per tail class. Doing so, we obtain the following experiment settings; {\bf i) CIFAR10\_Lt:} No. of classes 10, IF - 100, majority class frequency - 5,000, {\bf ii) CIFAR100\_Lt:} No. of classes 100, IF - 100, majority class frequency - 500, {\bf iii) ImageNet\_Lt:} No. of classes 1,000, IF - 256, majority class frequency - 1,280.

\subsection{Implementation}
Referring to the previous implementation, we use ResNet32 
 backbone for the CIFAR datasets and ResNext-50 
and ResNet50 
for the ImageNet\_Lt.
Models are trained with a batch size of 256. Learning rate for CIFAR10\_Lt and CIFAR100\_Lt are set to 0.2  and for ImageNet\_Lt 0.1. Similar to work of \cite{paco} and \cite{bcl}, we use Autoaugment (\cite{autoaug}) and Cutout (\cite{cutout}) for augmenting the features fed to the classifier. SimAugment (\cite{simaugment}) is used for the features fed to the contrastive loss. The model was trained on a Amazon Cloud VM with 8xV100 GPUs.

\subsection{Ablation Study}

We conducted an ablation study on CIFAR100\_Lt and CIFAR10\_Lt with IF set to 100 to analyse the performance benefits of RCL. The overall balanced accuracy (averaged per class) is reported in Table~\ref{t1} and Table~\ref{t2}, respectively. As large values heavily influence arithmetic means in obtaining the overall class accuracy, we also report the harmonic mean to reflect the performance 
of the worst-performing classes. Initially, the dataset is tested for cross-entropy loss with LA (\cite{menon2020long}). Where the logit adjustment parameter $\tau$ is set to 1.3 for RCL through grid search. Then, the accuracy is evaluated by combining standard supervised contrastive loss with setting  $\alpha$ to 2 and $\beta$ to 1 and using SimAugment for augmentation. The compression factor is set to 1 for the classes with validation accuracy less than $20\% $; for the other classes, the factor is set to 0.005. Compression is applied from the $100^{th}$ epoch.

CIFAR100\_Lt results in Table~\ref{t1} show adding RCL to standard SCL would increase top-1 accuracy by $1.1\%$ and harmonic mean accuracy by $3.2\%$ (rows 2 and 4). When RCL is added to BCL, the top-1 accuracy increases by $0.2\%$ and harmonic mean accuracy by $0.9\%$ (rows 3 and 5). SCL+RCL demonstrates the highest harmonic mean accuracy of all the combinations.

Similarly, Table~\ref{t2} shows the results for CIFAR10\_Lt where adding RCL to SCL  increases top-1 accuracy by $2.5\%$ and harmonic mean accuracy by $2.2\%$ (rows 2 and 4). When RCL is added to BCL, the top-1 accuracy increases by $1.9\%$ and harmonic mean accuracy by $2\%$ (rows 3 and 5). SCL+BCL+RCL demonstrates the highest harmonic mean accuracy of all.

\begin{figure}[bt]
    \centering
    \includegraphics[width=0.5\textwidth]{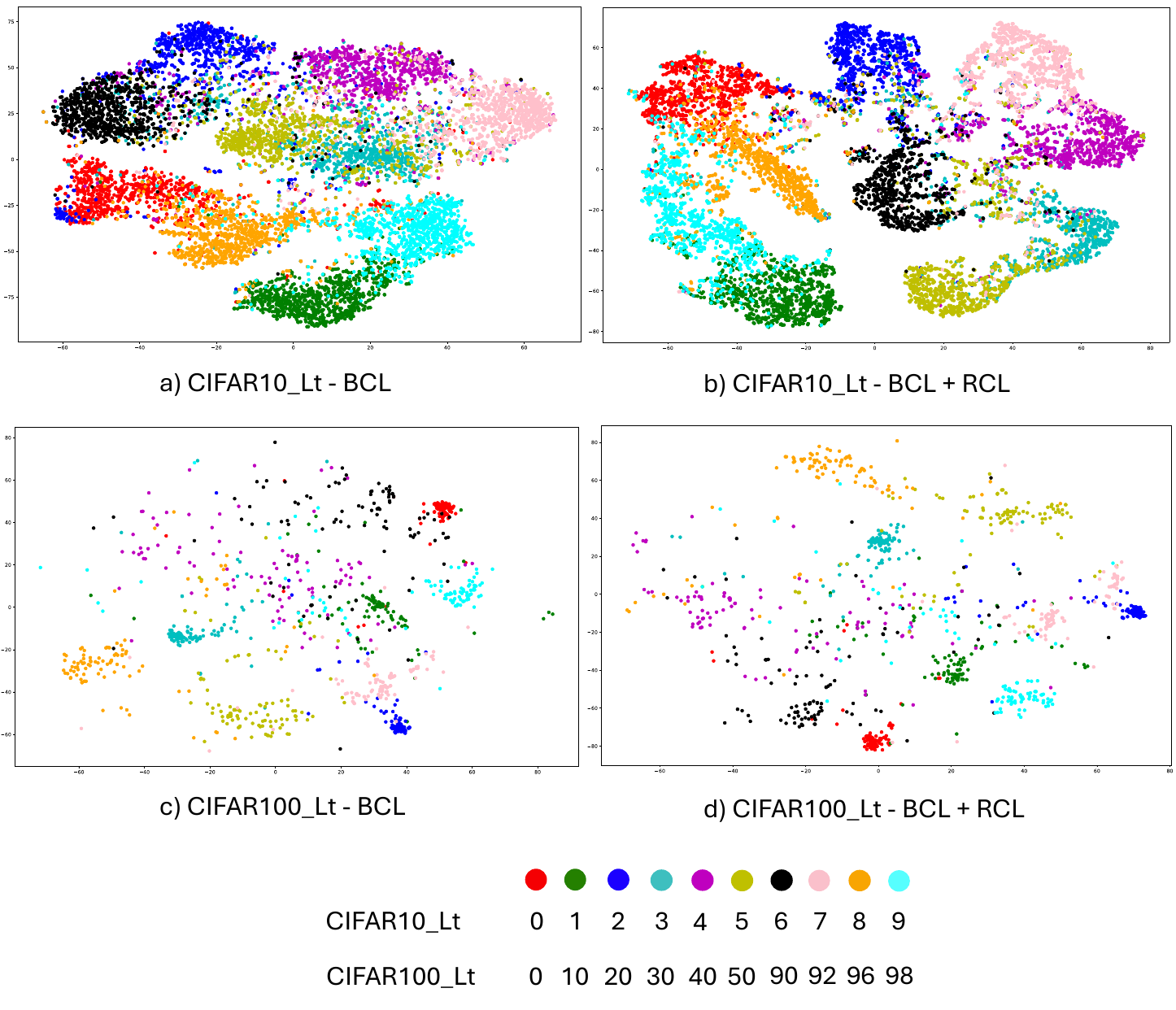}
    \caption{Learnt embedding distribution for CIFAR10\_Lt and CIFAR100\_Lt (IF=100:1) in TSNE format. Results showcase that the RCL addition has contributed to more balanced and separable tail class embeddings.} 
    \label{three}
    \vspace{-4mm}
\end{figure}

\subsection{Performance Analysis}

\subsubsection{Baselines}
We compare our proposed framework with 13 SOTA models under five main categories. Balanced overall test accuracy (arithmetic mean over per-class accuracy)
is used for comparison. The selected categories of algorithmic level solutions  are 1) Standard and modified (logit adjusted) CE minimization, 2) Two-stage learning methods where representation learning and classifier fine-tuning are handled in different stages; 3) Ensemble methods where multiple experts combined to enhance the head and tail class accuracy simultaneously, 4) CE loss combined with an additional loss to handle representation learning explicitly, 5) CE loss combined with SCL loss to enhance representation learning.

\subsubsection{Results}

We show the results in Table 4. As can be seen, standalone RCL demonstrates satisfactory performance compared to the SOTA methods. The improvement over BCL is 1.7\%,1.3\% and 0.6\% for CIFAR10\_Lt for IF = 100, 50 and 10, respectively. Similar to CIFAR100\_Lt, the values are 0.3\%,0.2\% and 0.7\%. Even though the CIFAR100\_Lt improvement over BCL is small compared to the other popular  SCL methods, RCL has a competitive performance.  Furthermore, BCL+RCL has consistent improvements 1.9\%, 1.2\%, and 0.9\%  for CIFAR10\_Lt. And for CIFAR100\_Lt, the corresponding improvements are  0.2\% for IF100 and 0.5 \% IF=10. In summary,  BCL performs better for CIFAR100\_Lt IF=50. For Imagenet\_Lt, BCL+RCL outperforms BCL by 0.1\% in both backbones according to Table~\ref{imagenet}.
\begin{table}[ht]
\centering
\caption{ Feature distribution evaluation based on class density and class separation using Calinski-Harabasz index (CHI) and Davies-Bouldin Index (DBI).  }
\begin{tabular}{ ccccc }

\hline
\textbf{Dataset (IF=100)}&\multicolumn{2}{c}{\textbf{BCL}}&\multicolumn{2}{c}{\textbf{BCL+RCL}}\\
\hline
&CHI &DBI & CHI&DBI \\
CIFAR100\_Lt &3,962.9&0.97&4,091.8 &0.94\\
CIFAR10\_Lt &84.1&4.1&105.8&4.0\\
\hline
\end{tabular}
\label{vmes}
\end{table}

\subsubsection{Embedding Space Analysis}

According to the quantitative analysis of the RCL method performance, it is evident that standalone RCL applied to the standard SCL demonstrates SOTA competitive accuracy levels. Apart from this, RCL demonstrates an interesting aspect when applied to the BCL. Even though the accuracy improvement over BCL is relatively small, adding the RCL to the BCL enhances the learnt feature distribution. In Figure~\ref{three}, we show the TSNE plots of the learnt representation of the model for CIFAR10\_Lt and CIFAR100\_Lt. BCL+RCL demonstrates a better class separation and balanced feature space division than BCL. As seen from Figure~\ref{three}(a), in BCL, tail class embeddings are grouped into sub-clusters and are far apart. Once the features compression is added, all embeddings corresponding to the same tail class are grouped together, providing better intra-class compactness and balancedness in the space.  Figure~\ref{three}(b) visually showcase the better class separation and balanced space distribution for all classes except for 1 and 9 compared to Figure~\ref{three}(a); furthermore, when Figures~\ref{three}(c) and (d) are compared, cluster balancedness has increased as a whole, as all classes occupy the feature space more equally compared to BCL when RCL is applied. In addition, compactness for classes 40 and 90 appears to increase significantly. To further validate this phenomenon, we compute the   Calinski–Harabasz index (\cite{calin}) for the learnt features of the validation set.  A higher index value indicates better cluster cohesion and cluster separation.  Table~\ref{vmes} indicates the index values for CIFAR10\_Lt and CIFAR100\_Lt.  Where we can see a consistent increment in the Calinski–Harabasz index for both datasets for BCL+RCL over BCL. Davies-Bouldin Index assesses the clustering quality where a lower value indicates more compact and well-separated clusters. According to the table, we can see that both datasets have a consistently lower value for BCL+RCL over BCL, which implies better cluster quality. The index is relatively insensitive to the choice of distance metric. Based on this analysis, we can conclude that RCL has the capability to optimize the distribution of the learnt feature simultaneously with the classification accuracy. The feature distribution is improved in balancedness, class compactness (cohesion) and regularization (larger margins, better class separation).

\section{Conclusion}
The paper proposed RCL for long-tail classification tasks. RCL addresses the balancedness of the learnt feature space,
enforces feature compactness and increases the margin width for tail classes. RCL shows a competitive level of accuracy 
over SOTA methods over three benchmark datasets with imbalance factors up to 256. RCL was also validated to perform well as a standalone and integrated loss on BCL, an existing supervised contrastive learning loss specifically designed for long-tail classification. It would be an important future work to investigate how much RCL can contribute to optimizing the learnt feature distribution with self-supervised contrastive learning under extreme class imbalance (i.e., IF $>$ than 256).

\section*{Acknowledgments}
This research was conducted by the University of Sydney for National Intelligence Postdoctoral Grant (project number NIPG-2022-006) and funded by the Australian Government.

\bibliographystyle{model5-names}
\bibliography{refs}

\end{document}